\documentclass{egpubl}
\SpecialIssueSubmission    

\usepackage[T1]{fontenc}
\usepackage{dfadobe}  

\usepackage{cite}  
\BibtexOrBiblatex
\electronicVersion
\PrintedOrElectronic
\ifpdf \usepackage[pdftex]{graphicx} \pdfcompresslevel=9
\else \usepackage[dvips]{graphicx} \fi

\usepackage{egweblnk}

\graphicspath{{}{pictures/}{images/}{./}} 

\usepackage{microtype}                 
\PassOptionsToPackage{warn}{textcomp}  
\usepackage{textcomp}                  
\usepackage{mathptmx}                  
\usepackage{times}                     
\usepackage{cite}                      
\usepackage{tabu}                      
\usepackage{booktabs}                  
\usepackage{multirow}
\usepackage{amsfonts}
\usepackage{newtxmath}
\usepackage{bigdelim}
\usepackage{bm}
\usepackage{wrapfig}

\title[A Task-driven Network for Mesh Classification and Semantic Part Segmentation]%
      {A Task-driven Network for \\ Mesh Classification and Semantic Part Segmentation}

\author[Dong et al.]{Qiujie Dong, Xiaoran Gong, Rui Xu, Zixiong Wang, Shuangmin Chen, Shiqing Xin, Changhe Tu, Wenping Wang \\ \textbf{Shandong University, Nankai University, Qingdao University of Science and Technology, Texas A$\&$M University}}  

\begin{document}

\teaser{
 \includegraphics[width=\linewidth]{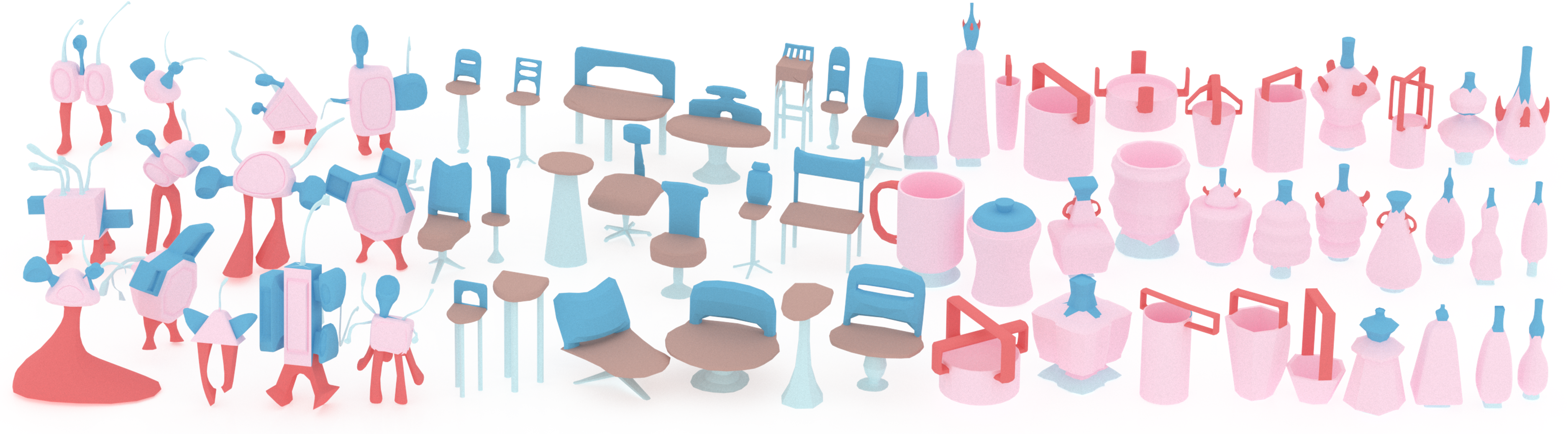}
 \centering
  \caption{A gallery of segmentation results of the COSEG dataset.
From left to right: Tele-aliens, Chairs, and Vases.
}
\label{fig:coseg}
}

\maketitle

\begin{abstract}

Given the rapid advancements in geometric deep-learning techniques, there has been a dedicated effort to create mesh-based convolutional operators that act as a link between irregular mesh structures and widely adopted backbone networks.
Despite the numerous advantages of Convolutional Neural Networks (CNNs) over Multi-Layer Perceptrons (MLPs), mesh-oriented CNNs often require intricate network architectures to tackle irregularities of a triangular mesh. These architectures not only demand that the mesh must be manifold and watertight but also impose constraints on the abundance of training samples.
In this paper, we note that for specific tasks such as mesh classification and semantic part segmentation, large-scale shape features play a pivotal role. This is in contrast to the realm of shape correspondence, where a comprehensive understanding of 3D shapes necessitates considering both local and global characteristics. Inspired by this key observation, we introduce a task-driven neural network architecture that seamlessly operates in an end-to-end fashion.
Our method takes as input mesh vertices equipped with the heat kernel signature (HKS) and dihedral angles between adjacent faces. Notably, we replace the conventional convolutional module, commonly found in ResNet architectures, with MLPs and incorporate Layer Normalization (LN) to facilitate layer-wise normalization.
Our approach, with a seemingly straightforward network architecture, demonstrates an accuracy advantage. It exhibits a marginal 0.1\% improvement in the mesh classification task and a substantial 1.8\% enhancement in the mesh part segmentation task compared to state-of-the-art methodologies. Moreover, as the number of training samples decreases to 1/50 or even 1/100, the accuracy advantage of our approach becomes more pronounced.
In summary, our convolution-free network is tailored for specific tasks relying on large-scale shape features and excels in the situation with a limited number of training samples, setting itself apart from state-of-the-art methodologies.


\end{abstract}

\section{Introduction}
Convolutional neural networks (CNNs), as underscored in existing literature~\cite{Redmon2016YouOL, Dosovitskiy2021AnII}, have attained substantial success in the realm of computer vision. This accomplishment has, in turn, ignited a growing interest in the field of geometric deep learning, focusing on the analysis of three-dimensional data.
In contrast to digital images, three-dimensional models can assume various representations, such as voxels~\cite{Klokov2017EscapeFC, Wang-adaptive-ocnn}, multi-view images~\cite{Su2015MVCNN, Le2017MVRNN}, point clouds~\cite{Qi2017PointNetDH, Qi2017PointNetDL, Li2018PointCNN}, and meshes~\cite{Hanocka2019MeshCNNAN, Lahav2020MeshWalkerDM}. Among these diverse representations, polygonal surfaces are particularly prevalent within the computer graphics community due to their capacity to provide precise topological and geometric information without inherent ambiguity.
However, it is crucial to acknowledge that polygonal surfaces differ from digital images in that they lack a uniform grid-based structure. This inherent irregularity presents significant challenges when employing CNNs for their analysis and processing.

There are two typical ways for handling mesh-based inputs. One way~\cite{Su2015MVCNN, Hanocka2019MeshCNNAN, He2020CurvaNetGD, Hu2022SubdivNet, Nicholas2022DiffusionNet, Dong2022Laplacian2MeshLM} involves regularizing mesh-based inputs to align them with the requirements of widely adopted deep convolutional networks~\cite{He2016ResNet, Ronneberger2015UNetCN, Hanocka2019MeshCNNAN}. For instance, techniques like DiffusionNet~\cite{Nicholas2022DiffusionNet} and Laplacian2Mesh~\cite{Dong2022Laplacian2MeshLM} transform the input mesh from the spatial domain into the spectral domain. In the spectral domain, an irregular mesh is typically represented as a combination of low-frequency signals.
The other way involves designing specialized modules to handle irregular triangles~\cite{Hanocka2019MeshCNNAN, Hu2022SubdivNet}. For example, MeshCNN~\cite{Hanocka2019MeshCNNAN} introduced a convolutional operator by encoding edge-centered features, leveraging the fact that each mesh edge is associated with four neighboring edges. On the other hand, SubdivNet~\cite{Hu2022SubdivNet} employs the loop subdivision algorithm to preprocess a coarse mesh. During the subdivision process, regular adjacency is established, enabling the definition of mesh-based convolutional kernels.
The underlying motivation for designing such intricate networks is to explore the potential to address a diverse range of analytical and processing tasks. However, attempting to use a single network for a wide variety of tasks is overly ambitious, considering the intrinsic disparities between these tasks.

In this paper, we observe that certain geometry analysis tasks heavily depend on large-scale shape features, while others require an understanding of shape at both local and global levels. Tasks like mesh classification and semantic part segmentation, for example, emphasize large-scale shape features, in contrast to the shape correspondence problem, which demands small-scale shape features.
Given this observation, it becomes imperative to design task-specific neural network architectures tailored to each respective task. Based on this fundamental insight, we propose an end-to-end neural network architecture specifically crafted to address mesh classification and semantic part segmentation tasks, both of which rely significantly on large-scale shape features.
In our implementation, we replace the typical convolutional module commonly found in ResNet architectures with Multi-Layer Perceptrons (MLPs) and incorporate Layer Normalization (LN) for layer-wise normalization. The network takes mesh vertices equipped with the heat kernel signature (HKS) and dihedral angles between adjacent faces as its input.

\begin{figure}[tb]
\begin{center}
 \includegraphics[width=\linewidth]{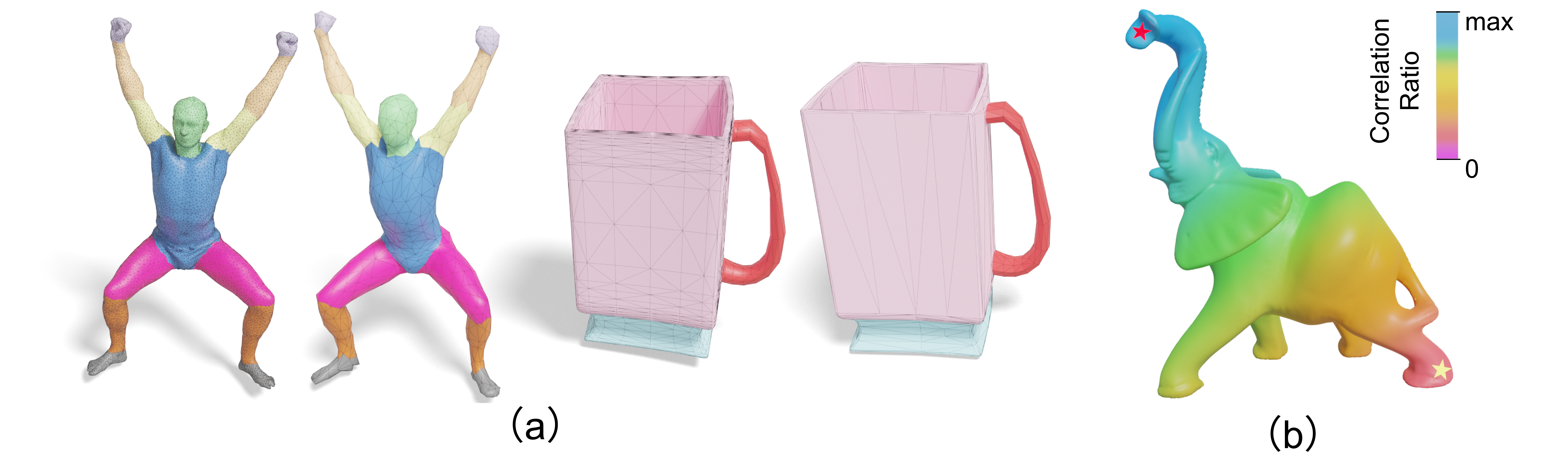}
\end{center}
  \caption{
  (a) Our method showcases a noteworthy ability to effectively handle mesh models with varying resolutions. Furthermore, it exhibits robustness when confronted with different triangulations, even in cases where the input model contains elongated and narrow triangles, as exemplified by the Vase model on the right.
  (b) We employ the red star as the reference for visualizing the correlation with other destination points. The color bar indicates that the weakest correlation is denoted by the color magenta. Remarkably, the correlation between the red star and the yellow star persists even when they are separated by a considerable distance.
  }
\label{fig:agnostic_globalInfluence}
\vspace{-1em}
\end{figure}

We conducted extensive experiments in shape classification and semantic part segmentation tasks. In contrast to meticulously crafted network architectures, our network, using the Residual Neural Network (ResNet)~\cite{He2016ResNet} as the foundational unit, eliminates the need for a convolution operator while demonstrating exceptional adaptability.
For instance, our network can handle meshes with arbitrary resolutions and triangulations, as illustrated in Figure~\ref{fig:agnostic_globalInfluence}(a). Moreover, it exhibits remarkable flexibility in accommodating a wide spectrum of geometry and topology complexities without compromising performance. Additionally, our task-driven network demonstrates a significant advantage when dealing with a limited number of training samples.

Our contributions are summarized as follows:
\begin{itemize}
\item We introduce a task-driven neural network that accommodates diverse mesh triangulations/resolutions and manifold/non-manifold mesh models.

\item Our method demonstrates robust learning capabilities for mesh classification and semantic segmentation tasks, surpassing existing state-of-the-art methods, especially in scenarios with limited data availability.

\item Extensive experiments were conducted, and all the results consistently validate the efficacy of our proposed approach.
\end{itemize}

\section{Related Work}

Our research falls within the domain of geometric deep learning~\cite{Bronstein2021GeometricDL}. Consequently, in this section, we begin with an overview of diverse shape representations employed in the field of geometric deep learning for characterizing 3D shapes. Following this, we delve into an examination of mesh-based neural networks, specifically designed to tackle the inherent irregularities associated with polygonal surfaces. For a more in-depth exploration of related works, please refer to~\cite{Xiao2020survey, RN242survey}.

\subsection{3D Shape Deep Learning}
Various forms of representation have been employed as inputs in the realm of geometric deep learning, including voxels, point clouds, multi-view images, and triangle meshes. Adapting neural networks designed for digital images to work with voxel representations is generally straightforward. For instance, Wu et al.~\cite{Wu20153DSA} introduced a 3D convolutional network tailored for voxel representations, primarily for shape classification and retrieval tasks. Due to the structured nature of 3D voxels, voxel representations have become a prevalent choice, especially in applications like shape reconstruction. To handle the escalating complexity associated with voxels, octree-based volumetric representations~\cite{Wang2017OCNN, Riegler2017OctNetLD} have gained increased attention.

Point clouds have gained prominence, particularly in the context of autonomous driving. Notable methods such as PointNet~\cite{Qi2017PointNetDL} and PointNet++~\cite{Qi2017PointNetDH} have been developed to extract features from point clouds by establishing spatial correlations among points through K-nearest neighbor (KNN) search. PointCNN~\cite{Li2018PointCNN} introduces a $\chi$-transformation to reorder input points into a latent and potentially canonical order, while PCT~\cite{Guo2021PCTPC} leverages transformer mechanisms to enhance overall performance. Despite these advancements, point clouds still struggle to accurately encode geometry and topology, limiting their utility in tasks based on polygonal surfaces.

Another category involves encoding a 3D shape using a collection of 2D images, which has garnered significant attention. A common method to achieve this is by projecting a 3D shape from multiple views~\cite{Su2015MultiviewCN, Sinha2016DeepL3} using rendering techniques or from a panoramic view~\cite{Shi2015DeepPanoDP} through cylinder projection around its principal axis. The 3D-to-2D transformation can also be accomplished through parametrization. For example, Sinha et al.~\cite{Sinha2016DeepL3} proposed using authalic parametrization on a spherical domain to generate geometry images, allowing standard CNNs to directly learn 3D shapes. However, despite their ease of implementation, these approaches tend to struggle when encoding complex geometry or topology.

\subsection{Mesh Based Deep Learning}
Numerous strategies have been proposed to address the irregularities inherent in polygonal surfaces. In addition to the approach of encoding a 3D shape using a collection of 2D images, three prevalent strategies stand out.

The first strategy involves transforming irregular connections into a more regular organization. For instance, MeshCNN~\cite{Hanocka2019MeshCNNAN} formulates convolutions based on the correlation of each edge in the mesh with four other edges. It employs an edge collapse operation to define pooling operations that preserve surface topology. In contrast, SubdivNet~\cite{Hu2022SubdivNet} is a CNN framework designed for 3D triangle meshes with a Loop subdivision sequence connectivity. It leverages the hierarchical multi-resolution structure provided by a subdivision surface, where each face in a closed 2-manifold triangle mesh is adjacent to exactly three faces. However, it's important to note that these approaches directly operate on triangles, which can make it challenging to ensure consistent outputs for different triangulations of the same object.

The second strategy centers around the use of graph neural networks (GNNs). In these approaches, meshes are transformed into graph structures, where vertices correspond to nodes within the graph, and edges represent connections between these vertices. Notable methods like DGCNN~\cite{Wang2019DGCNN} and MDGCNN~\cite{Adrien2018MDGCNN} adopt a graph-based representation to model three-dimensional meshes. MeshWalker~\cite{Lahav2020MeshWalkerDM} employs random walks to explore the mesh topology, enabling the encoding of localized features. However, it comes with a substantial computational cost, particularly as the mesh resolution increases. On the other hand, PD-MeshNet~\cite{Milano2020PrimalDualMC} encodes adjacency between faces through both primal and dual graphs, with the pooling operation grounded in the mesh simplification algorithm. CurvaNet~\cite{He2020CurvaNetGD}, meanwhile, utilizes graph neural networks to handle polygonal surfaces, learning geometric features from a differential geometry perspective.

The last strategy, gaining popularity recently, involves using the Laplacian operator to implicitly represent the geometry and topology of meshes. Leveraging eigendecomposition within a differentiable pipeline, HodgeNet~\cite{Smirnov2021HodgeNetLS} proposes a spectral-based learning technique. DiffusionNet~\cite{Nicholas2022DiffusionNet} and Laplacian2Mesh~\cite{Dong2022Laplacian2MeshLM} employ spectral-domain transformations to handle irregular meshes. The former utilizes spectral acceleration techniques to diffuse vertex features for local-to-global information accumulation, while the latter employs spectral reconstruction techniques to transform meshes into multi-resolution spectral spaces. It's worth noting that these approaches, due to the sensitivity of the Laplacian operator to triangulation, may yield varying outputs for different triangulations.

\section{Experiments}
In our research, we conducted comprehensive experiments to assess the effectiveness of our approach in both mesh classification and semantic part segmentation tasks. To maintain consistency, all meshes were scaled to a unit scale before training. During the training process, data augmentation was applied by randomly selecting one of the Euler angles, specifically 0, $\pi$/2, or $\pi$, and then performing rotations along the $x$-axis, $y$-axis, or $z$-axis. This augmentation technique helps diversify the training data and enhance the model's robustness.

\begin{table}[t]
  \caption{
  The table presents classification accuracy statistics for the SHREC11 dataset, which comprises two versions: one containing simplified mesh models~\cite{Hanocka2019MeshCNNAN} and the other containing dense mesh datasets~\cite{Lian2011SHRECT}. Each version of the dataset is evaluated using two Train-Test Split plans: split-16, with 16 models for training, and split-10, with 10 models for training. In the table, the use of bold fonts indicates the best performance. The ``-S'' suffix denotes experiments conducted on simplified mesh models~\cite{Hanocka2019MeshCNNAN}, while the ``-D'' suffix indicates experiments conducted on dense mesh datasets~\cite{Lian2011SHRECT}.
  }
  \label{tab:shrec11}
\begin{center}
  \setlength{\tabcolsep}{1mm}{
  \begin{tabular}{%
	l%
	*{2}{c}%
	}
  \toprule
  \textbf{Method} & \textbf{Split-16} & \textbf{Split-10}\\
  \midrule
  GWCNN~\cite{Ezuz2017GWCNNAM} & 96.6$\%$ & 90.3$\%$\\
MeshCNN~\cite{Hanocka2019MeshCNNAN} & 98.6$\%$ & 91.0$\%$\\
PD-MeshNet~\cite{Milano2020PrimalDualMC} & 99.7$\%$ & 99.1$\%$ \\
MeshWalker~\cite{Lahav2020MeshWalkerDM} & 98.6$\%$ & 97.1$\%$  \\ 
SubdivNet~\cite{Hu2022SubdivNet} & 99.9$\%$ & 99.5$\%$\\
HodgeNet~\cite{Smirnov2021HodgeNetLS} & 99.2$\%$ & 94.7$\%$\\
DiffusionNet~\cite{Nicholas2022DiffusionNet} & $-$ & 99.5$\%$\\
FC~\cite{Mitchel2021FC} & $-$ & 99.2$\%$\\
HSN~\cite{Wiersma2020HSN} & $-$ & 96.1$\%$\\
Ours -S & \textbf{100}$\%$ & \textbf{99.7}$\%$ \\
Ours -D & \textbf{100}$\%$ & \textbf{100}$\%$ \\
  \bottomrule
  \end{tabular}}%
  \end{center}
  \vspace{-2em}
\end{table}

\subsection{Mesh Classification}
\textbf{SHREC11.}
For the mesh classification experiment, we employed the SHREC-11 dataset~\cite{Lian2011SHRECT}, which consists of 30 classes, each with 20 examples. To ensure consistency, we followed the evaluation setup proposed by~\cite{Ezuz2017GWCNNAM} and conducted evaluations based on the 10-10 and 16-4 train-test splits. The original dataset~\cite{Lian2011SHRECT} includes 3D shapes with over 18,000 faces in the mesh, while the simplified dataset~\cite{Hanocka2019MeshCNNAN} contains 3D shapes with only 500 faces.

In our evaluation, we adhered to the approach used in related studies and tested our method on both the reduced and original SHREC-11 datasets. The results, as presented in Table~\ref{tab:shrec11}, demonstrate that our method outperforms other approaches on both train-test splits. Specifically, on the reduced SHREC-11 dataset, our method achieves 100\% classification accuracy on split-16 and 99.7\% on split-10. On the original SHREC-11 dataset, our method attains 100\% accuracy on both split-16 and split-10. These results highlight the effectiveness and robustness of our approach in mesh classification tasks.

\textbf{Manifold40.}
The Manifold40 dataset~\cite{Hu2022SubdivNet} represents a larger and more challenging dataset, encompassing 12,311 CAD models spanning 40 different categories. This dataset was reconstructed from ModelNet40~\cite{Wu20153DSA}, with a primary objective of improving the triangulation quality of the meshes. However, the reconstruction and simplification operations carried out during the dataset construction process introduce slight shape differences and significant variations in tessellation when compared to the original models.

In Table~\ref{tab:manifold40}, we present a performance comparison of our method with other approaches on the Manifold40 dataset~\cite{Hu2022SubdivNet}. Our method achieves performance that is comparable to SubdivNet~\cite{Hu2022SubdivNet} and surpasses other methods in terms of accuracy. Significantly, our approach eliminates the need for the intricate mesh subdivision step~\cite{Loop1987SmoothSS}, which is a part of SubdivNet~\cite{Hu2022SubdivNet}. This outcome underscores the effectiveness of our approach in handling large and demanding datasets like Manifold40, delivering competitive performance without requiring additional complex operations.

\begin{table}[ht]
  \caption{The classification accuracy statistics on Manifold40~\cite{Hu2022SubdivNet}.}
  \label{tab:manifold40}
\begin{center}
  \begin{tabular}{%
	l%
	c%
    c%
	}
  \toprule
  \textbf{Method} & \textbf{Input} & \textbf{Accuracy} \\
  \midrule
  PointNet++~\cite{Qi2017PointNetDH}        & point cloud & 87.9$\%$ \\
  PCT~\cite{Guo2021PCTPC}                   & point cloud & \textbf{92.4}$\%$ \\
  \midrule
  MeshNet~\cite{feng2019meshnet}            & mesh & 88.4$\%$ \\
  MeshWalker~\cite{Lahav2020MeshWalkerDM}   & mesh & 90.5$\%$ \\
  Laplacian2Mesh~\cite{Dong2022Laplacian2MeshLM}  & mesh & 90.9$\%$ \\
  SubdivNet~\cite{Hu2022SubdivNet}          & mesh & \textbf{91.2}$\%$ \\
  Ours  & mesh & 90.9$\%$ \\
  \bottomrule
  \end{tabular}%
  \end{center}
\end{table}

\subsection{Mesh Semantic Part Segmentation}
For the evaluation of our method in semantic part segmentation tasks, we utilized three distinct datasets: COSEG dataset~\cite{Wang2012ActiveCO}, human body dataset~\cite{Maron2017ConvolutionalNN}, and the IntrA dataset~\cite{yang2020intra}. The primary aim of semantic part segmentation is to predict per-face segmentation labels for the meshes. In our evaluation, we adopted ``hard'' labels~\cite{Milano2020PrimalDualMC} as the evaluation metric. These labels measure face-wise accuracy, which is different from ``soft'' labels~\cite{Hanocka2019MeshCNNAN}, which are based on edge-wise accuracy.

\textbf{COSEG.}
The COSEG dataset is comprised of three distinct shape datasets: Chairs, Vases, and Tele-aliens, which contain 400, 300, and 200 triangle meshes, respectively. The Vases and Tele-aliens datasets have meshes labeled into 4 parts, while the Chairs dataset has meshes labeled into 3 parts. Similar to the approach taken in MeshCNN~\cite{Hanocka2019MeshCNNAN}, we randomly split the dataset into 85\% training and 15\% testing sets for evaluation.

The quantitative results of our method on the Chairs, Vases, and Tele-aliens datasets are presented in Table~\ref{tab:coseg}. Our method demonstrates state-of-the-art performance on the Chairs and Vases datasets and delivers comparable results on the Tele-aliens dataset. It is noteworthy that the Tele-aliens dataset contains a substantial number of long and narrow triangles, which particularly benefits methods like MeshCNN~\cite{Hanocka2019MeshCNNAN} and HodgeNet~\cite{Smirnov2021HodgeNetLS}, as they utilize per-edge features as part of their input. The visualized results of semantic part segmentation can be observed in Figure~\ref{fig:coseg}. Notably, the results reported by SubdivNet~\cite{Hu2022SubdivNet} involve mapping the prediction results back to the raw meshes. To maintain fairness in our comparison, we did not include SubdivNet in our evaluation of the semantic part segmentation task.

\begin{table}[tb]
  \caption{The mesh segmentation accuracy statistics on the COSEG dataset~\cite{Wang2012ActiveCO}.}
  \label{tab:coseg}
\begin{center}
  \setlength{\tabcolsep}{1mm}{
  \begin{tabular}{%
	l%
	*{3}{c}%
	}
  \toprule
  \textbf{Method} & \textbf{Chairs} & \textbf{Vases} & \textbf{Tele-aliens} \\
  \midrule
  PointNet~\cite{Qi2017PointNetDL} & 70.2$\%$ & 91.5$\%$  & 54.4$\%$ \\
  DCN~\cite{Xu2017DCN}  & 95.7$\%$ & 90.9$\%$  & - \\
  MeshCNN~\cite{Hanocka2019MeshCNNAN}   & 93.0$\%$ & 92.4$\%$  & \textbf{96.3}$\%$ \\
  HodgeNet~\cite{Smirnov2021HodgeNetLS} & 95.7$\%$ & 90.3$\%$  & 96.0$\%$ \\
  Laplacian2Mesh~\cite{Dong2022Laplacian2MeshLM} & 96.6$\%$ & 94.6$\%$  & 95.0$\%$ \\
  Ours & \textbf{97.5}$\%$ & \textbf{95.1}$\%$  & 95.9$\%$ \\
  \bottomrule
  \end{tabular}}%
  \end{center}
\end{table}

\textbf{Human Body Segmentation.}
We further extend our evaluation to include the human body dataset, which encompasses 370 training 3D shapes from various sources such as Adobe Fuse, FAUST~\cite{Bogo2014FAUSTDA}, MIT~\cite{Vlasic2008ArticulatedMA}, and SCAPE~\cite{Klokov2017EscapeFC}. The test set comprises 18 3D shapes from the SHREC07~\cite{giorgi2007shape} humans dataset. These meshes have all been meticulously labeled by Maron et al.~\cite{Maron2017ConvolutionalNN} and manually segmented into 8 parts.
In line with most approaches, including ours, we employ the version of this dataset processed by MeshCNN~\cite{Hanocka2019MeshCNNAN}, which downsamples each mesh to 1500 faces. This dataset serves as a widely accepted benchmark for evaluating semantic part segmentation performance specifically on human body shapes.

\begin{figure}[t]
 \begin{center}
 \includegraphics[width=0.8\linewidth]{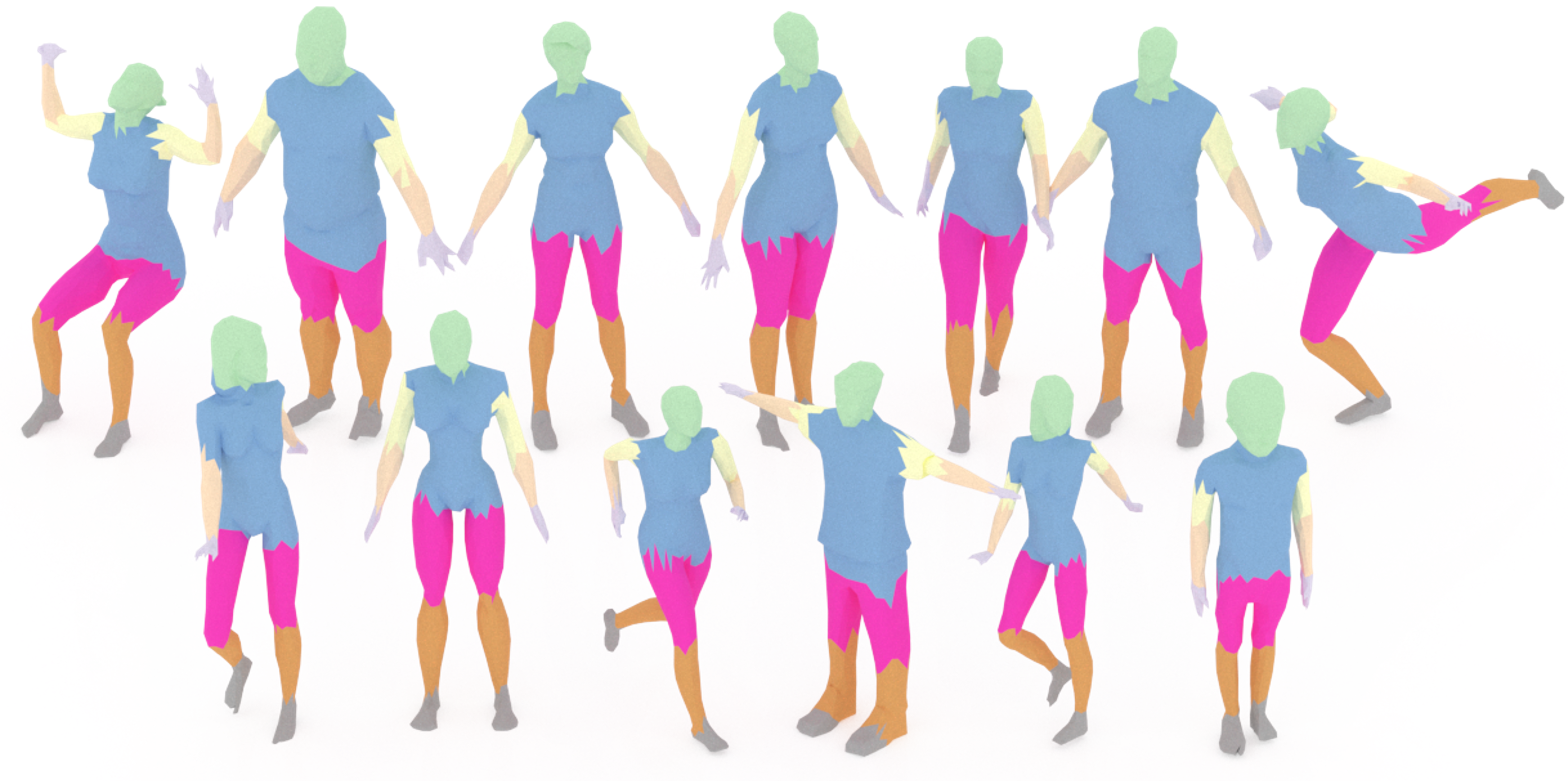}
 \end{center}
 \caption{We conducted tests on the Human Body dataset.
 The segmentation results are visually depicted in this figure.
 }
 \label{fig:humanbody_seg}
\end{figure}

\begin{table}[tb]
  \caption{The mesh segmentation accuracy statistics on the Human-Body dataset~\cite{Maron2017ConvolutionalNN}. The options ``xyz'' and ``hks'' correspond to the raw coordinates and the heat kernel signatures, respectively.}
  \label{tab:humanbody}
\begin{center}
  \setlength{\tabcolsep}{1mm}{
  \begin{tabular}{%
	l%
	*{2}{c}%
	}
  \toprule
  \textbf{Method} & \textbf{Input} & \textbf{Accuracy}\\
  \midrule
  PointNet~\cite{Qi2017PointNetDL}  & point cloud & 74.7$\%$\\
  PointNet++~\cite{Qi2017PointNetDH} & point cloud & 82.3$\%$\\
  \midrule
  GCNN~\cite{Boscaini2015GCNN} & mesh & 85.9$\%$ \\
  MDGCNN~\cite{Adrien2018MDGCNN} & mesh & 87.2$\%$\\
  MeshCNN~\cite{Hanocka2019MeshCNNAN} & mesh & 85.4$\%$\\
  PD-MeshNet~\cite{Milano2020PrimalDualMC} & mesh & 85.6$\%$\\
  HodgeNet~\cite{Smirnov2021HodgeNetLS} & mesh & 85.0$\%$\\
  DiffusionNet (xyz)~\cite{Nicholas2022DiffusionNet} & mesh & 88.8$\%$ \\
  DiffusionNet (hks)~\cite{Nicholas2022DiffusionNet} & mesh & 90.5$\%$ \\
  Laplacian2Mesh~\cite{Dong2022Laplacian2MeshLM} & mesh & 88.6$\%$\\
  Ours & mesh & \textbf{90.6}$\%$\\
  \bottomrule
  \end{tabular}}%
  \end{center}
\end{table}

\begin{table*}[tb]
\centering
\caption{The mesh segmentation accuracy statistics on the IntrA dataset~\cite{yang2020intra}.
In this context, ``vessel'' refers to parent vessel segments, and ``aneurysm'' pertains to aneurysm segments.}
\label{tab:intra}
\begin{tabular}{lccccc} 
\toprule
\multirow{2}{*}{\textbf{Method}} & \multicolumn{2}{c}{\textbf{IoU}} & \multicolumn{3}{c}{\textbf{DSC}}  \\ 
\cmidrule{2-3} 
\cmidrule{5-6}
   & \multicolumn{1}{l}{vessel} & \multicolumn{1}{l}{aneurysm} & & vessel & aneurysm \\
\cmidrule{1-6}
PointNet~\cite{Qi2017PointNetDL} & 74.0$\%$ & 37.3$\%$ & & 84.1$\%$ & 49.0$\%$\\
PointNet++~\cite{Qi2017PointNetDH} & 93.4$\%$ & 76.2$\%$ & & 96.5$\%$ & 83.9$\%$\\
PointCNN~\cite{Li2018PointCNN} & 92.5$\%$ & 70.7$\%$  & & 96.0$\%$ & 78.6$\%$\\
SO-Net~\cite{li2018so} & 94.2$\%$ & 80.1$\%$  & & 97.0$\%$ & 87.9$\%$\\ 
SpiderCNN~\cite{xu2018spidercnn} & 90.2$\%$ & 67.3$\%$ & & 94.5$\%$ & 75.8$\%$\\
PointConv~\cite{wu2019pointconv} & 94.2$\%$ & 79.1$\%$ & & 96.9$\%$ & 86.0$\%$\\
GS-Net~\cite{xu2020geometry} & 90.1$\%$ & 64.5$\%$ & & 94.6$\%$ & 74.5$\%$\\
PCT~\cite{Guo2021PCTPC} & 92.5$\%$ & 78.1$\%$ & & 96.1$\%$ & 85.8$\%$\\
AdaptConv~\cite{zhou2021adaptive} & 90.5$\%$ & 70.3$\%$ & & 96.0$\%$ & 80.6$\%$\\
PAConv~\cite{xu2021paconv} & 92.0$\%$ & 78.7$\%$ & & 95.7$\%$ & 87.6$\%$\\
3DMedPT~\cite{yu20213d} & 94.8$\%$ & 81.8$\%$ & & 97.3$\%$ & 89.3$\%$\\ 
Niemann et al.~\cite{Niemann2023DeepLS} & 74.8$\%$ & 71.4$\%$ & & - & -\\
Laplacian2Mesh~\cite{Dong2022Laplacian2MeshLM} & 90.2$\%$ & 75.1$\%$ & & 91.5$\%$ & 89.9$\%$\\ 
Ours & \textbf{95.2}$\%$ &\textbf{87.7}$\%$ & & \textbf{97.5}$\%$ & \textbf{93.0}$\%$\\
\bottomrule
\end{tabular}
\end{table*}

Table~\ref{tab:humanbody} provides a comprehensive overview of the accuracy statistics for various methods assessed on the human body dataset. Notably, our approach achieves performance that is comparable to DiffusionNet~\cite{Nicholas2022DiffusionNet} and surpasses other methods in the dataset. It is noteworthy that several of the compared methods require intricate preprocessing steps, such as re-meshing triangle meshes~\cite{Hanocka2019MeshCNNAN} and constructing primal-dual graphs~\cite{Milano2020PrimalDualMC}.

Furthermore, our approach outperforms MDGCNN~\cite{Adrien2018MDGCNN}, which employs dynamic graph convolution. The qualitative results presented in Figure~\ref{fig:humanbody_seg} demonstrate the efficacy of our method in accurately and consistently segmenting parts of the human body.

\textbf{IntrA.}
Our evaluation includes a medical dataset, specifically the IntrA dataset, which comprises 116 annotated samples for the binary part segmentation task. Medical experts manually annotate the aneurysm segments in this dataset, with the scale of each aneurysm segment determined by the requirements of preoperative examination.

\begin{figure}[tb]
 \begin{center}
 \includegraphics[width=0.8\linewidth]{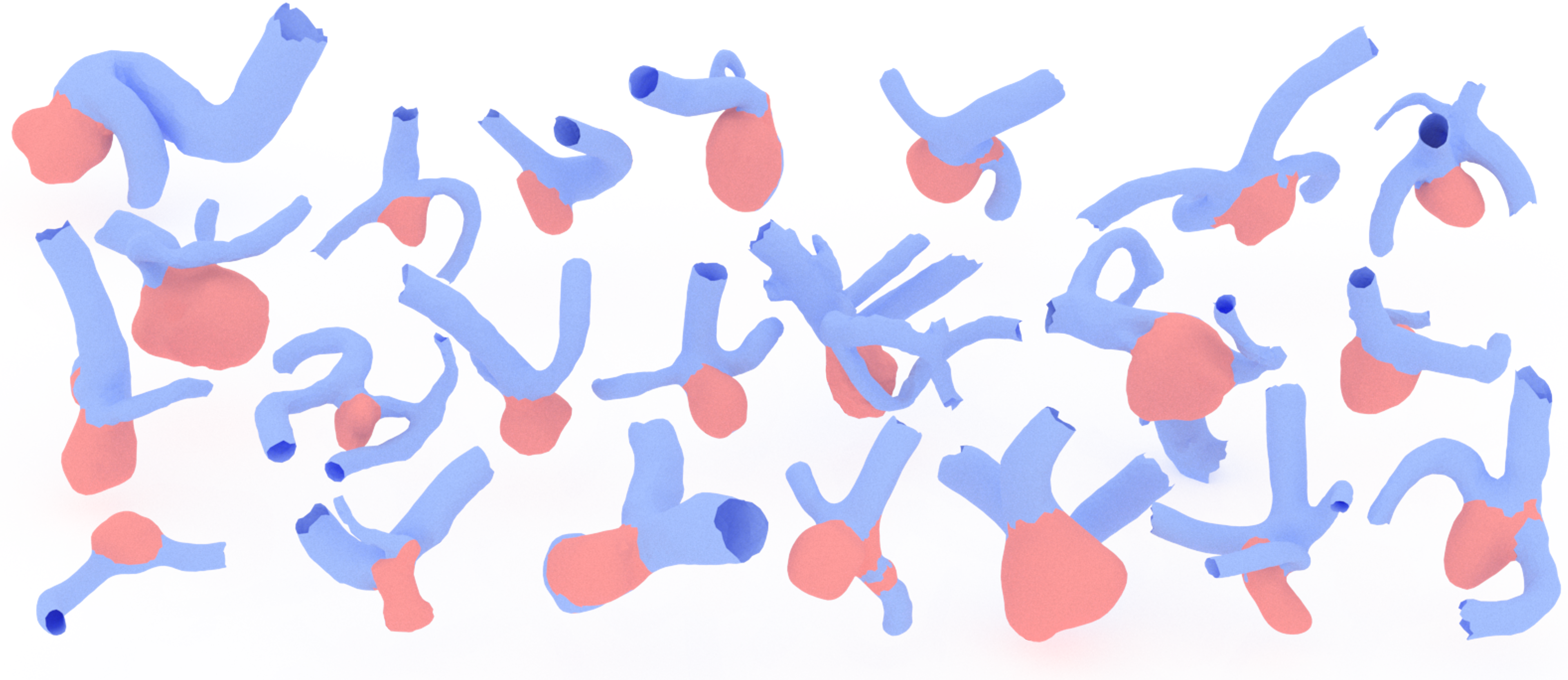}
 \end{center}
 \caption{The segmentation results for intracranial aneurysms demonstrate the high accuracy achieved by our method.
 }
 \label{fig:blood_seg}
\end{figure}

Table~\ref{tab:intra} presents the performance evaluation using Point Intersection over Union (IoU) and the Sørensen–Dice coefficient (DSC) as evaluation metrics. Notably, our method achieves the best results for parent vessel segmentation in terms of both IoU and DSC, surpassing other methods that also use 512 input points.
Figure~\ref{fig:blood_seg} further provides a qualitative evaluation of our method on medical data, demonstrating its superiority in segmenting medical images.

\subsection{Small Training Sample Size}
Numerous contemporary approaches~\cite{Hu2022SubdivNet, Dong2022Laplacian2MeshLM} heavily depend on the amount of training samples. However, in most scenarios, there is a limited amount of data available. Our task-driven network design mitigates the reliance on a large number of labeled data for effective training.

To validate this, we randomly choose 1/10, 1/50, and 1/100 (rounded up) of the data from the original training set to train our part segmentation network, while the size of the data in the test set remains unchanged. Table~\ref{tab:lean_on_data} illustrates that despite a significant reduction in training data, our approach maintains substantial prediction accuracy.

For instance, in the human body dataset where shapes have a similar structure, even with the training data reduced to as little as 1/100 of the original size (i.e., only 4 labeled human meshes), our segmentation accuracy amounts to 85.7\%. For datasets like chairs, vases, tele-aliens, and Manifold40, despite the shape diversity, 1/100 of the training data (4 chair meshes, 3 vase meshes, 2 tele-aliens meshes, 3 meshes for each class in Manifold40, respectively) still yields considerable segmentation/classification accuracy. This showcases high accuracy in scenarios with limited data availability.

\renewcommand\arraystretch{1.1}
\begin{table*}[h]
\caption{The mesh part segmentation accuracy or classification accuracy statistic on different datasets with fewer data.
In split datasets (Humanbody, Chairs, Vases, Tele-aliens), 'N' represents the total number of data samples in the original training set.
Meanwhile, in classification datasets (Manifold40) comprising 40 distinct categories, 'N' signifies the count of data samples within each category present in the original dataset.
}
\label{tab:lean_on_data}
\begin{center}
\setlength{\tabcolsep}{3mm}{
\begin{tabular}{c|lllll}
\hline
\textbf{Dataset} & \multicolumn{1}{c}{\textbf{Methods}} & \multicolumn{1}{c}{\textbf{N}} & \multicolumn{1}{c}{$\lceil$\textbf{N/10}$\rceil$} & \multicolumn{1}{c}{$\lceil$\textbf{N/50}$\rceil$} & \multicolumn{1}{c}{$\lceil$\textbf{N/100}$\rceil$} \\ \hline
\multirow{2}{*}{HumanBody (N=381)}
& Laplacian2mesh~\cite{Dong2022Laplacian2MeshLM} &  88.6$\%$  &  70.1$\%$  &   61.5$\%$  & 56.3$\%$ \\
& Ours &  \textbf{90.6}$\%$ &  \textbf{88.8}$\%$  & \textbf{87.5}$\%$  &  \textbf{85.7}$\%$ \\ 
\hline
\multirow{2}{*}{Chairs (N=337)}
& Laplacian2mesh~\cite{Dong2022Laplacian2MeshLM} & 96.6$\%$ & 68.5$\%$ & 60.6$\%$ & 45.9$\%$ \\
& Ours  &  \textbf{97.5}$\%$ & \textbf{88.7}$\%$ & \textbf{83.2}$\%$  & \textbf{67.5}$\%$ \\ 
\hline
\multirow{2}{*}{Vases (N=252)}
& Laplacian2mesh~\cite{Dong2022Laplacian2MeshLM}   & 94.6$\%$  & 74.3$\%$  &   49.8$\%$ & 40.5$\%$ \\
& Ours  &  \textbf{95.1}$\%$ & \textbf{87.1}$\%$ & \textbf{78.9}$\%$  & \textbf{70.1}$\%$ \\ 
\hline
\multirow{2}{*}{Tele-aliens (N=169)}
& Laplacian2mesh~\cite{Dong2022Laplacian2MeshLM}  & 95.0$\%$ & 75.2$\%$  &   50.2$\%$ & 25.6$\%$ \\
& Ours  & \textbf{95.9}$\%$ & \textbf{86.9}$\%$  & \textbf{80.2}$\%$ & \textbf{59.6}$\%$ \\
\hline
\multirow{2}{*}{Manifold40 (N=246)}
& Laplacian2mesh~\cite{Dong2022Laplacian2MeshLM}  & 90.9$\%$ & 56.3$\%$  &   35.6$\%$ &  20.2$\%$ \\
& Ours  & \textbf{90.9}$\%$ & \textbf{76.5}$\%$  & \textbf{55.6}$\%$ & \textbf{47.5}$\%$ \\
\hline
\end{tabular}}
\end{center}
\end{table*}

\begin{figure}[h]
\begin{center}
 \includegraphics[width=0.8\linewidth]{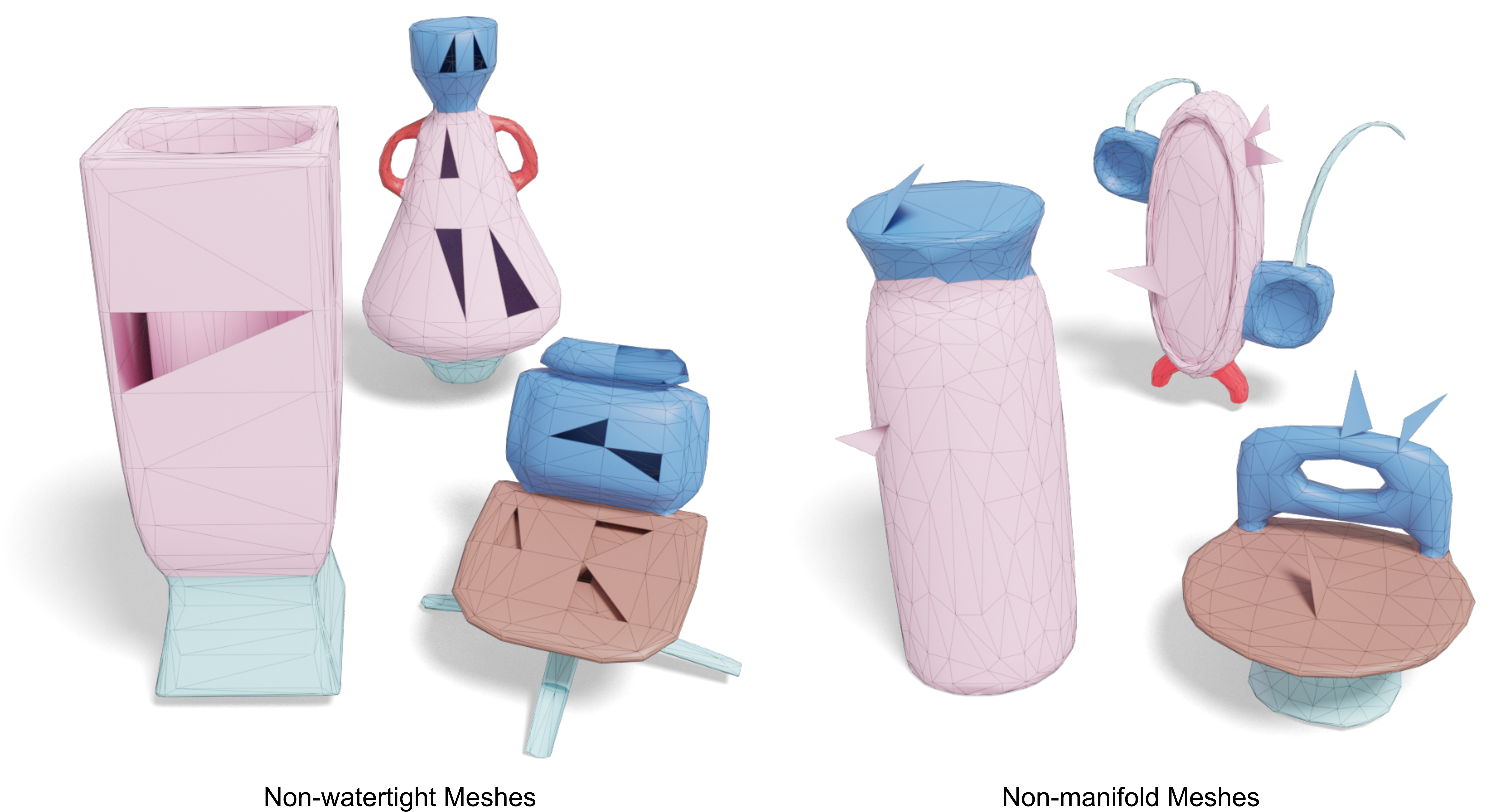}
\end{center}
  \caption{
Our method demonstrates the capability to perform segmentation on non-watertight data, as shown on the left, and non-manifold data, as depicted on the right.
  }
\label{fig:nonmanifold_nonwatertight}
\end{figure}

\subsection{Robustness and Efficiency}
Numerous existing mesh segmentation methods, such as SubdivNet~\cite{Hu2022SubdivNet} and MeshCNN~\cite{Hanocka2019MeshCNNAN}, are designed with the assumption that the input mesh must be watertight, manifold, or both. Unfortunately, this limitation narrows their applicability to a more restricted range of mesh datasets.

In stark contrast, our method approach leverages vertices as the medium for features, allowing us to seamlessly handle both manifold and non-manifold mesh models (as illustrated in Figure~\ref{fig:nonmanifold_nonwatertight}). Moreover, the strategic use of a batch size of 1 and gradient accumulation techniques empowers our network to effectively process meshes with varying resolutions and triangulations (as depicted in Figure~\ref{fig:agnostic_globalInfluence}(a)). This exceptional flexibility and adaptability make our approach well-suited for tackling diverse mesh structures and scenarios.

We conducted performance tests of our model on the human body dataset using an experimental platform equipped with a 24G RTX 3090 GPU and a dual-core AMD EPYC 7642 CPU. In the pre-computation stage, which involves calculating the Heat Kernel Signature (HKS), the average time required is 305ms. It's worth noting that this pre-calculation is a one-time process for the entire dataset. For the training and testing stages, the average times are 90ms and 20ms, respectively. These results underscore the efficiency of our model in processing mesh data for both training and inference tasks, further enhancing its practical utility.

\section{Limitations}
Our method harnesses the Heat Kernel Signature (HKS) as a valuable tool for preserving essential topological information within mesh data. The HKS serves as a representation of geometric characteristics that capture vital aspects of the mesh's structure. However, a limitation becomes apparent when dealing with disconnected objects, 
heat cannot flow from one component to another, making the HKS struggle to effectively encode the local-to-global geometric features~\cite{Nicholas2022DiffusionNet}.

Additionally, MLP-based methodologies
establish a global receptive field by connecting all neurons, unlike convolutional neural networks (CNNs) that use local connectivity between neurons. In CNNs, each neuron is connected only to nearby neurons in the next layer, allowing them to capture local features efficiently.
Therefore, our method does not exhibit superiority on those local-feature based tasks, such as shape correspondence.
To evaluate our method's performance on the shape correspondence task, we conducted a train-test split of 8:2 on the FAUST~\cite{Bogo2014FAUSTDA} and SCAPE~\cite{Anguelov2005SCAPESC} datasets, following a training procedure similar to~\cite{Hu2022SubdivNet, Mehta2017VNectR3}. Subsequently, we computed 30-dimensional functional coordinates and constructed a functional map between the source and target meshes by solving a linear system, following a procedure akin to that in~\cite{ovsjanikov2012functional}.
The results of our method on the shape correspondence task, as presented in Table~\ref{tab:correspondence}, demonstrate that our approach achieves comparable results to other methods but does not surpass them.

\begin{table}[tb]
  \caption{Shape Correspondence Error Comparison. `F' and `S' indicate FAUST~\cite{Bogo2014FAUSTDA} and SCAPE~\cite{Anguelov2005SCAPESC} datasets. `F on S' means training on FAUST and testing on SCAPE and vice versa.}
  \label{tab:correspondence}
  \begin{center}
  \setlength{\tabcolsep}{2mm}{
  \begin{tabular}{%
	l%
	*{4}{r}%
	}
  \toprule
  \textbf{Method} & \textbf{F} & \textbf{S} & \textbf{F on S} & \textbf{S on F}\\
  \midrule
  BCICP~\cite{BCICP} & 15.$\%$ & 16.$\%$ & - & - \\
  ZoomOut~\cite{Zoomout} & 6.1$\%$ & 7.5$\%$ & - & - \\
  SURFMNet~\cite{SURFMNet} & 7.4$\%$ & 6.1$\%$ & 19.$\%$ & 23.$\%$ \\
  FMNet~\cite{FMNet} & 5.9$\%$ & 6.3$\%$ & 11.$\%$ & 14$\%$ \\
  3D-CODED~\cite{3D-CODED} & 2.5$\%$ & 31.$\%$ & 31.$\%$ & 33.$\%$ \\
  GeomFMaps~\cite{GeomFMaps} & \textbf{1.9}$\%$ & \textbf{3.0}$\%$ & \textbf{9.2}$\%$ & 4.3$\%$ \\
  SubdivNet~\cite{Hu2022SubdivNet} & \textbf{1.9}$\%$ & \textbf{3.0}$\%$ & 10.5$\%$ & \textbf{2.6}$\%$ \\
  ours & 6.9$\%$ & 7.2$\%$ & 16.8$\%$ & 13.2$\%$ \\
  \bottomrule
  \end{tabular}}%
  \end{center}
  \vspace{-2.5em}
\end{table}

\section{Conclusion and Future Work}
\textbf{Conclusion.}
In this research, we introduced a task-based neural network for mesh classification and semantic part segmentation tasks. Our network architecture shares similarities with ResNet but distinguishes itself by replacing all convolution operations with simpler MLPs. Additionally, we employ Layer Normalization (LN) for effective normalization.

Our network offers several key advantages, including the elimination of complex and time-consuming preprocessing steps traditionally associated with input data. Furthermore, it circumvents the need for specialized network architectures tailored to address the irregularities inherent in mesh structures. Through extensive experiments, we have empirically demonstrated that our method, despite its remarkable simplicity, consistently achieves competitive results when compared to existing methods. These results underscore the effectiveness and efficiency of our approach within the domain of mesh-based tasks, particularly in scenarios where data availability is restricted.

\textbf{Future Work.}
The simplicity and effectiveness of our method make it an attractive option for researchers and practitioners in the field. 
There are several promising research directions for further development.
One avenue for future research is to conduct further fine-tuning of the network to enhance its performance in shape classification and segmentation tasks. This could involve exploring different network architectures, normalization techniques, or loss functions to push the boundaries of its capabilities in these specific tasks.
Another research direction is to expand our network into a more comprehensive and versatile framework that can handle a broad spectrum of mesh-related tasks and applications. This may involve developing additional headblocks and adapting the network's architecture to address tasks dependent on local information, which could include tasks like mesh denoising, deformation prediction, or shape completion. Finally, as mesh data can be large and complex, optimizing our method for efficiency and scalability is crucial. Future work could explore techniques to accelerate training and inference times while maintaining or even improving performance.


\section{Architecture of our Neural Network}

\begin{figure}[htb]
\begin{center}
 \includegraphics[width=\linewidth]{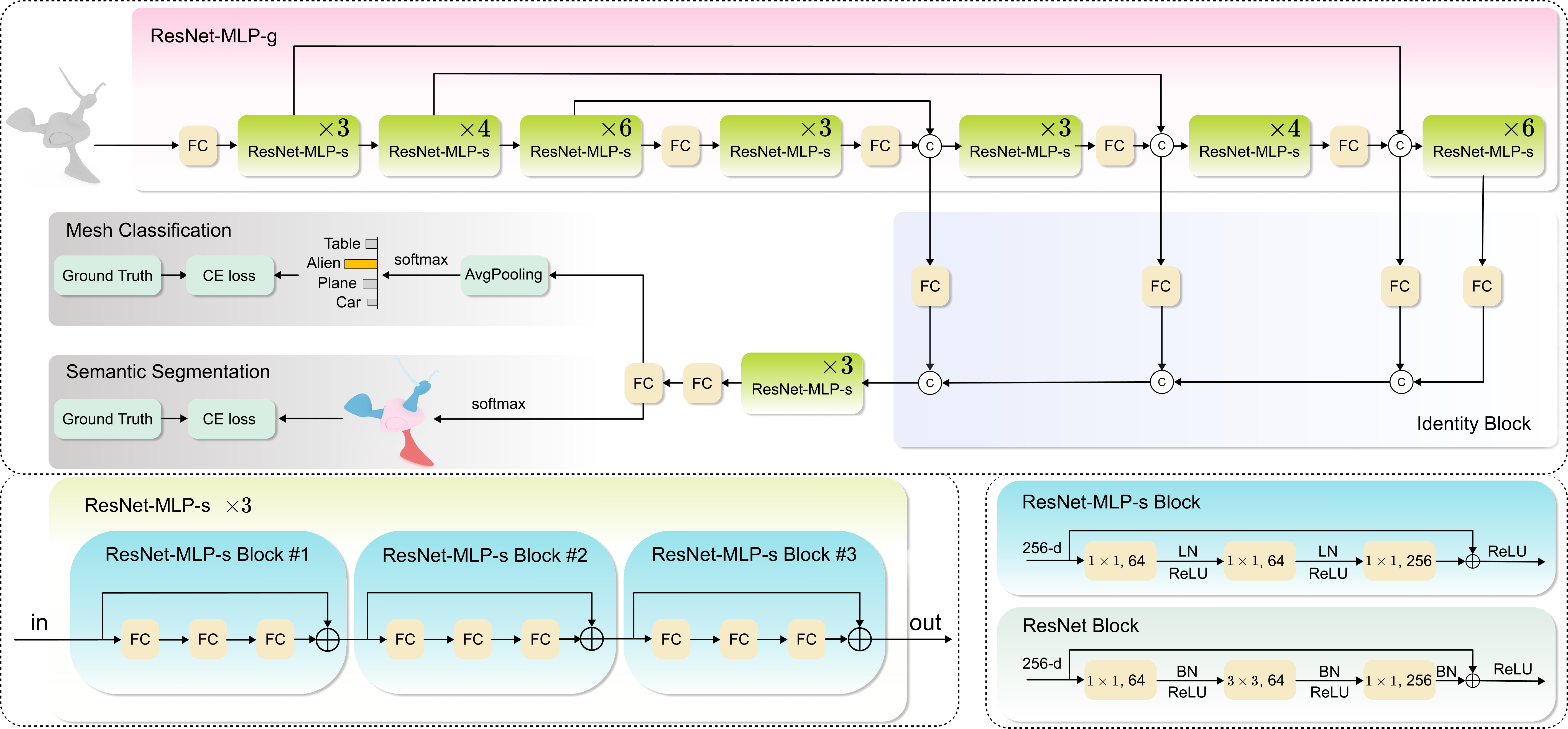}
\end{center}
  \caption{
  An overview of our comprehensive network designed for tackling mesh classification and segmentation tasks. It incorporates a bottleneck structure to efficiently reduce computational requirements. The circled ``C'' and circled ``+'' symbols indicate the concatenation and sum operations, respectively.
It's important to highlight that we've replaced the conventional convolutional layer in the original ResNet with an MLP layer, denoted as ResNet-MLP-g. Layer normalization (LN) is applied to normalize the intermediate outputs within this layer. The ResNet-MLP-g is composed of several ResNet-MLP-s modules, and each ResNet-MLP-s module consists of multiple fully connected (FC) layers. To illustrate the difference between our ResNet-MLP-s and the traditional ResNet block, we've provided a visual comparison in the lower right part of the figure.
  }
\label{fig:pipeline}
\end{figure}

In Figure~\ref{fig:pipeline}, we elaborate on our streamlined network architecture.
Our network comprises multiple ResNet-like blocks, utilizing fully connected layers across all network layers, where the specific parameters of each layer are shown in Table~\ref{tab:architecture}.
Throughout the training process, we apply layer normalization to the data and subsequently utilize the ReLU nonlinear activation function. Additionally, for tasks like classification and semantic part segmentation, we access distinct head blocks within the network to execute specific tasks.

\newcommand{\blockb}[3]{\multirow{3}{*}{\(\left[\begin{array}{c}\text{1$\times$1, #1}\\[-.1em] \text{1$\times$1, #2}\\[-.1em] \text{1$\times$1, #2}\end{array}\right]\)$\times$#3}
}

\begin{table}[h]
  \caption{
We provide detailed size information regarding the building blocks within our network architecture.
}
  \label{tab:architecture}
\begin{center}
\scriptsize 
  \begin{tabular}{c|c|c}
  \toprule
  \textbf{Layer Name} & \textbf{Output Size} & \textbf{Layer Architecture}\\
  \hline
    & 256 & 1$\times$1, input\_size \\
  \hline
  \multirow{3}{*}{ResNet-MLP-s \#1} & \multirow{3}{*}{256} & \blockb{256}{64}{3}\\
  &  &\\
  &  &\\
  \hline
  \multirow{3}{*}{ResNet-MLP-s \#2} & \multirow{3}{*}{256} & \blockb{256}{64}{4}\\
  &  &\\
  &  &\\
  \hline
  \multirow{3}{*}{ResNet-MLP-s \#3} & \multirow{3}{*}{256} & \blockb{256}{64}{6}\\
  &  &\\
  &  &\\
  \hline
  & 512 & 1$\times$1, 256 \\
  \hline
  \multirow{3}{*}{ResNet-MLP-s \#4} & \multirow{3}{*}{512} & \blockb{512}{128}{3}\\
  &  &\\
  &  &\\
  \hline
  & 256 & 1$\times$1, 512 \\
  \hline
  \multirow{3}{*}{ResNet-MLP-s \#5} & \multirow{3}{*}{512} & \blockb{512}{128}{3}\\
  &  &\\
  &  &\\
  \hline
  & 256 & 1$\times$1, 512 \\
  \hline
  \multirow{3}{*}{ResNet-MLP-s \#6} & \multirow{3}{*}{512} & \blockb{512}{128}{4}\\
  &  &\\
  &  &\\
  \hline
  & 256 & 1$\times$1, 512 \\
  \hline
  \multirow{3}{*}{ResNet-MLP-s \#7} & \multirow{3}{*}{512} & \blockb{512}{128}{6}\\
  &  &\\
  &  &\\
  \hline
  \multirow{3}{*}{ResNet-MLP-s \#8} & \multirow{3}{*}{512} & \blockb{512}{128}{3}\\
  &  &\\
  &  &\\
  \hline
  & 128 & 1$\times$1, 512 \\
  \hline
  & 32 & 1$\times$1, 128 \\
  \hline
  \multicolumn{3}{c}{Head Blocks for Different Tasks} \\
  \bottomrule
  \end{tabular}%
  \end{center}
\end{table}

\section{Ablation Studies}
In our study, we perform ablation experiments on multiple aspects, including network structure, input features, and normalization techniques. These experiments allow us to analyze and quantify the importance of each component and make informed decisions about the model's design.

\textbf{Convolution Patterns.}
Network design is a critical aspect of deep learning, and it plays a pivotal role in determining the performance of a model. In our study, we carefully considered various network architectures and design choices to achieve the best results. Table~\ref{tab:ConvoPatterns} outlines the progression of our network architecture, starting with fully convolutional layers (Conv) and gradually incorporating modern design modules such as ResNet (Conv + Res). However, the combination of MLP with residual learning (MLP + Res) ultimately yielded the best results in our experiments.

It's worth noting that we initially explored using pure MLP layers for our tasks. Unfortunately, this configuration did not produce satisfactory results. The primary challenge we encountered was the occurrence of a severe gradient disappearance phenomenon within the network architecture, which led to a significant drop in overall network performance. This highlights the importance of choosing an appropriate network design that can effectively address such issues and improve the model's performance.

\textbf{Input Features.}
Table~\ref{tab:InputFeatures} showcases the outcomes of our experiments, illustrating the effectiveness of various input features for each vertex in our network. One key observation is the pivotal role played by the Heat Kernel Signature (HKS) in our tasks. 
As a critical component to encode both local-to-global shape variations,
its inclusion substantially enhances the network's performance, equipping it with the capacity to effectively process and comprehend intricate mesh structures in the context of mesh classification and semantic part segmentation tasks. This demonstrates the value of carefully selecting and incorporating relevant input features to empower the network to handle the complexities of 3D mesh data.

\textbf{Normalization.}
Normalization techniques hold substantial sway over the performance of neural network architectures. In our research, we undertook a reevaluation of the effects of various normalization techniques in the context of our tasks, as succinctly outlined in Table~\ref{tab:Normalization}. A noteworthy observation is that employing Layer Normalization (LN) directly within the original ResNet architecture might not yield optimal results, aligning with prior findings~\cite{GN2018arXiv180308494W}.

However, in the case of our method, we discerned a different trend. LN emerged as the most fitting and effective normalization method for our specific network architecture. It delivered superior performance in our mesh classification and semantic part segmentation tasks, underscoring its suitability for the demands of our 3D mesh data processing. This highlights the importance of adapting normalization techniques to the unique characteristics and requirements of the task at hand, and in our case, LN proved to be the optimal choice.

\begin{table}[t]
  \caption{The mesh segmentation accuracy statistics for different network structures.}
  \label{tab:ConvoPatterns}
\begin{center}
  \setlength{\tabcolsep}{1mm}{
  \begin{tabular}{%
	lccc
	}
  \toprule
  \textbf{Backbone} & \textbf{Human-Body} & \textbf{Tele-aliens} & \textbf{Chairs}\\
  \midrule
  Conv &  84.0$\%$ & 94.0$\%$ & 96.6$\%$ \\
  Conv + Res & 88.2$\%$ & 94.1$\%$ & 96.7$\%$ \\
  MLP + Res (Ours) &  \textbf{90.6}$\%$ & \textbf{95.9}$\%$ & \textbf{97.5}$\%$ \\
  MLP  & 25.2$\%$ & 31.3$\%$ & 35.8$\%$ \\
  \bottomrule
  \end{tabular}}%
  \end{center}
\end{table}

\begin{table}[h]
  \caption{The mesh segmentation accuracy statistics for different input features.}
  \label{tab:InputFeatures}
\begin{center}
  \setlength{\tabcolsep}{1mm}{
  \begin{tabular}{%
	l%
	*{3}{c}%
	}
  \toprule
  \textbf{Input Features} & \textbf{Human-Body} & \textbf{Tele-aliens} & \textbf{Manifold40}\\
  \midrule
  Vertex coordinates & 70.3$\%$ & 65.0$\%$ & 72.3$\%$ \\
  + Vertex normal & 76.5$\%$ & 77.4$\%$ & 75.1$\%$ \\
  + dihedral angle & 83.1$\%$ & 87.1$\%$ & 81.2$\%$ \\
  + HKS (full) & \textbf{90.6}$\%$ & \textbf{95.9}$\%$ & \textbf{90.9}$\%$ \\
  \bottomrule
  \end{tabular}}%
  \end{center}
\end{table}

\begin{table}[h]
  \caption{The mesh segmentation accuracy statistics for different Normalizations.}
  \label{tab:Normalization}
\begin{center}
  \setlength{\tabcolsep}{1mm}{
  \begin{tabular}{%
	l%
	*{2}{c}%
	}
  \toprule
  \textbf{Normalization} & \textbf{Human-Body} & \textbf{Chairs} \\
  \midrule
  Batch Norm~\cite{Ioffe2017BatchRT} & 85.4$\%$ & 85.8$\%$ \\
  Group Norm~\cite{GN2018arXiv180308494W} & 87.0$\%$ & 96.9$\%$ \\
  Global Response Norm~\cite{GRN2023arXiv230100808W} & 87.0$\%$ & 95.5$\%$ \\
  Instance Norm~\cite{Ulyanov2016InstanceNT} & 88.5$\%$ & 96.9$\%$ \\
  Layer Norm~\cite{Ba2016LayerN} & \textbf{90.6}$\%$ & \textbf{97.5}$\%$ \\
  \bottomrule
  \end{tabular}}%
  \end{center}
  \vspace{20em}
\end{table}

\bibliographystyle{eg-alpha-doi}

\bibliography{egbibsample}

\end{document}